Clustering for Improved Learning in Maze Traversal Problem

Undergraduate Honors Thesis

The University of Memphis

In partial fulfillment of

Bachelor of Science in Electrical Engineering

and

Bachelor of Science in Computer Engineering

Eddie White

May 2009

Abstract

The maze traversal problem (finding the shortest distance to the goal from any position in a maze) has been an interesting challenge in computational intelligence. Recent work has shown that the cellular simultaneous recurrent neural network (CSRN) can solve this problem for simple mazes. This thesis focuses on exploiting relevant information about the maze to improve learning and decrease the training time for the CSRN to solve mazes. Appropriate variables are identified to create useful clusters using relevant information. The CSRN was next modified to allow for an additional external input. With this additional input, several methods were tested and results show that clustering the mazes improves the overall learning of the traversal problem for the CSRN.

Table of Contents

1. Introduction

      Maze navigation is an important subject for autonomous robot and route optimization. An example application may be finding the optimized route in an urban area during a disaster. Teaching the robot to navigate through an unknown environment and find the optimal route is a very difficult task. A simplified version of this problem can be simulated by using a random 2D synthetic maze. This research seeks to improve the performance of an existing cellular simultaneous recurrent neural network (CSRN) for the 2D maze navigation problem by clustering the states of the maze. For this problem, this maze is represented as a square matrix. Within this square matrix, each cell is classified as an obstacle, pathway, or the goal. Figure 1 shows an illustration of a maze.

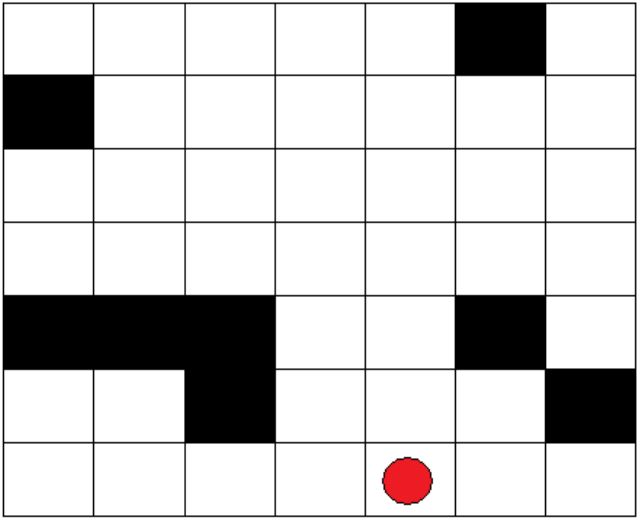

*Figure 1: Illustration of 2D Maze Navigation Problem.*

      The 2D maze representation has already been explored in several works[1][2][3][4][5]. The problem has already been shown to have promising results using a CSRN trained with a

back propagation algorithm [1]. Further work replaced the back propagation through time training algorithm with the extended Kalman filter (EKF) [2]. The maze traversal performance in [2] is good but still has room for improvement. Additionally, it has been noted [3] that the CSRN fails to converge at times. Experiments with the CSRN show that non-convergence of the CSRN occurs more often when using larger amounts of data. This work seeks to improve the accuracy and speed of the network by clustering the maze cells.

During the summer of 2008, members of the Intelligent Systems and Image Processing (ISIP) group at the University of Memphis visited the Information and Communication Technology (ICT) lab at the University of Wollongong, Australia on a National Science Foundation sponsored face recognition project. It was there that the idea of clustering mazes was first conceived. One of the students at the ICT lab was working on training algorithms for neural networks that used weighted cluster information to help learn pattern classification of large or imbalanced data sets [6]. The success of weighted clusters in [6] inspired us to use clusters to improve the performance of the CSRN for maze navigation. Since the problem domain of maze traversal is inherently different then classification, investigation was required to find a plausible clustering technique.

Literature research yielded a relevant work in which Manor et al. [7] used clustering to improve the learning in a Q learning problem. In [7], the basic idea was to use partial information learned about an environment to accelerate learning. The steps include first interacting with the environment and learning using Q-learning, saving the state transition history, and if the clustering conditions are met then the state transition history is translated into a graph representation. The clustering algorithm is then run, new options are learned for reaching neighborhood clusters from each cluster, and finally the new options are added to the agent's choices. An option is basically a set of state transitions to take. So, if an agent knows that it needs to get to a certain location and there is an option leading to that location then the agent can simply take that option instead of each individual state transition leading to the desired location. The timing for the clustering is very important. If clustering is performed too soon the agent will not have enough useful information to make any improvements by clustering. However, if clustering is performed too late then the agent has already learned most of the environment and will not have a need for many of the options. Two clustering techniques were used. The first is clustering by topology and the second is clustering by value [7]. When clustering is performed by topology, their results were that each cluster was basically a different room.

2. Background

2.1 Neural Networks: The Neuron

The human brain is very powerful and complex. For years, scientists and researchers have worked to discover how the brain works in hopes of emulating the learning and calculations brains are capable of. The brain consists of several processing units called neurons. When a message is sent between neurons, it is sent by way of the synapses or connecting links. There are an estimated $10^{10}$ neurons connected by approximately $10^{14}$ synapses in the human brain [8]. Each synapse has an associated weight that either amplifies or diminishes the signal passing through it. Each neuron typically receives several inputs which it sums together. The sum is then passed through an activation or squashing function before being sent to other neurons [9]. Figure 2 shows an illustration of the processing of a single neuron.

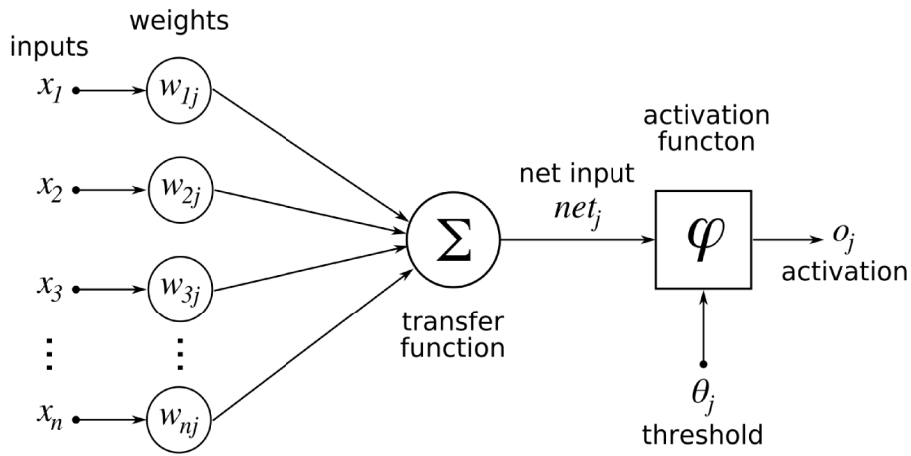

*Figure 2: Neuron [4].*

2.2 Neural Network History

In 1943 McCulloch and Pitts introduced the idea of neural networks as computing machines. Later in 1949, Hebb postulated the first rule for self-organizing, or unsupervised, learning. Unsupervised learning takes place based on a set of rules rather than examples. The network performs calculations and updates weight in attempts to best apply the defined rules to its input data. In 1958, Rosenblatt proposed the perceptron as the first model for supervised learning [9]. Supervised learning presents the network with the desired output after it has performed its calculations. Based off the error between the network's calculated output and the desired output, it adjusts it weights and tries again. This is basically teaching by example. The network is presented with several example data, training data, for which it tries to emulate the desired output, target. By connecting several of these perceptons in certain arrangements, it was shown that this artificial neural network could be taught simple functions.

Artificial neural networks learn by adjusting weights based off an error. The network is trained by first being presented with an input for which it calculates an output. For supervised learning, this output is then compared to a target output. An error is then calculated based upon the difference between the output of the network and the target output. This error is then used to adjust the weights in the network to better approximate the target output. Throughout this process, the total error decreases and the network is considered to have converged once the total error does not vary greatly for several iterations. This means that the network has adjusted the weights to the best of its ability and in doing so, it has minimized the error. Some of the first functions taught to neural networks were the AND and OR functions.

2.3 Types of Neural Networks

It was quickly discovered that the basic perceptron had limited utility. Work was then done to allow the neural network to learn more complex functions. In general, three types of network structures have evolved; single layer, multilayer, and recurrent. Single layer networks simply have an input layer and an output layer. This configuration is considered a single layer network because no calculations are done during the input layer. Single layer networks are considered strictly feed forward. Multilayer networks have 1 of more hidden layers. Hidden layers are the layers between the input and output layers. By adding multiple layers, there are more connections and thus, more weights which allow for more learning. Obviously an artificial neural network cannot approach the number of neurons and connections that are in the human brain, but more neurons and layers can added to help solve more complex problems. Adding layers enables the network to solve higher order statistics [9]. More layers are particularly useful when the network takes in a large number of inputs, has a large input layer.

Recurrent networks differ from other networks because of the arrangement of some of their connections. Many networks are fully connected which means that each neuron is connected to every neuron in the layer before and after it. Certain network configurations may not have every neuron connected to every other neuron in its neighboring layers. This network would be considered partially connected. Recurrent networks differ by having nodes from one layer feed into nodes from a previous layer. Basically, outputs of certain nodes are used in the next iteration for some other nodes in the network. Feedback can results in non-linear dynamics behavior if the neural network contains non-linear units [9].

One example of a recurrent network that has been developed is the simultaneous recurrent network (SRN). Previous research has shown that functions generated by multi-layer perceptrons (MLPs) are always able to be learned by SRNs, but the opposite is not true [1]. SRNs, being recurrent networks, use the output of the current iteration as input for the next iteration. This makes them excellent for prediction problems and training with the Kalman filter, since they are able to learn from previous estimations. The basic topology of the SRN is shown in Fig. 3. The output is calculated by multiplying the previous output (z) and the input (x) by the appropriate weights (W).

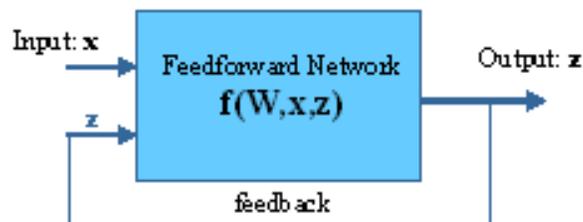

*Figure 3: The Basic Topology of a SRN.*

Cellular neural networks (CNN) consist of identical elements, arranged in some sort of geometry as shown in Fig. 4. Due to the symmetry of the network, each element is able to share the same weights. Decreasing the number of weights can significantly decrease the time needed to train the network. The symmetry of cellular neural networks can also be useful in solving problems that contain a similar sort of inherent geometry, such as the maze navigation problem. Each element of such a network can be as simple as an artificial neuron or more complex, as a MLP.

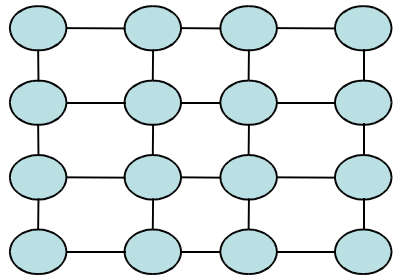

*Figure 4: A typical CNN architecture.*

The CSRN is a combination of a CNN and a SRN. The idea of the CSRN is biologically motivated. The behavior of the CSRN imitates the cortex of the brain which consists of columns similar to each other. The CSRN was, at one time, trained with back propagation through time (BPTT). However, BPTT is very slow. In [1], the extended Kalman filter (EKF) is implemented to train the network by state estimation. The structure of the network is shown in Fig. 5. The CSRN is structured so that there are two external input nodes (obstacle and goal), four neighbor nodes, and five recurrent nodes. The number of recurrent nodes is defined by a variable within the code that is easily changed.

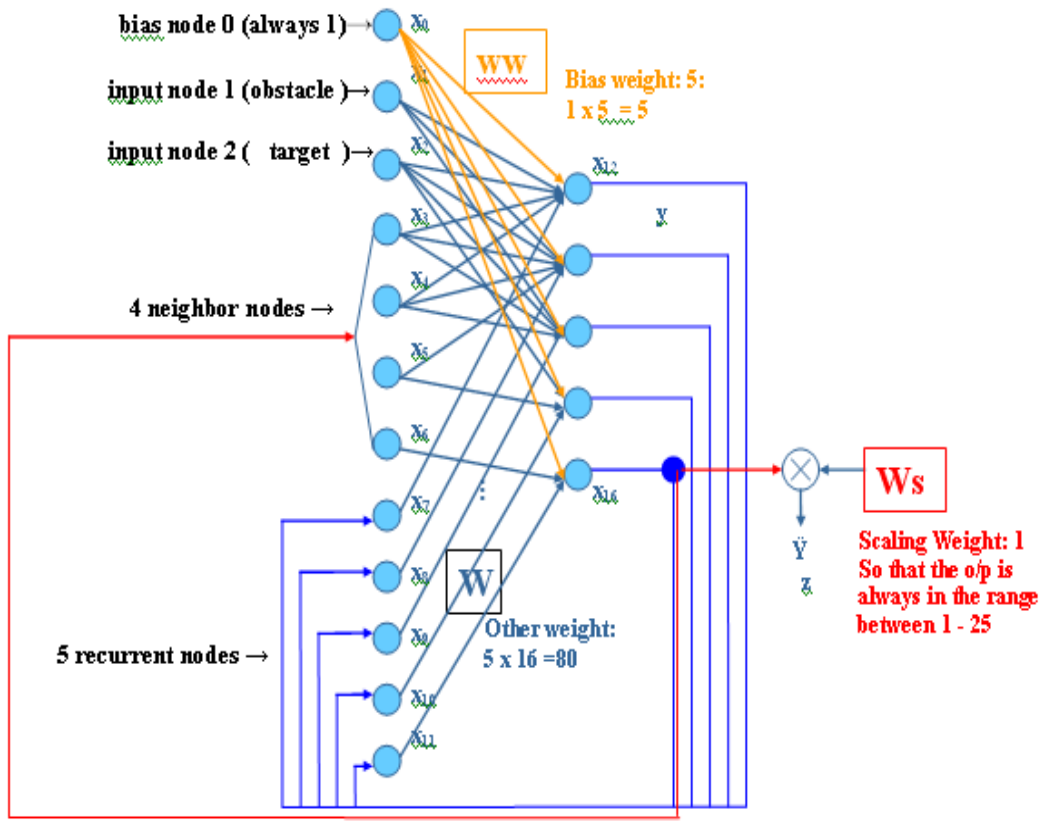

*Figure 5: Network Structure of the CSRN.*

2.4 Learning in Neural Network

Neural networks learn by adjusting the weights between neurons. This is why adding neurons and connections can, sometimes, help a network to learn more efficiently. The adjustments to the weights are calculated as the derivative of the error with respect to the particular weight. Once the network has been trained and the error has been minimized, it is then tested. At this point the network is simply presented with data and calculates an output. If it outputs the correct solution then the network has successfully learned the given task. Often, a network may learn the correct output for some data but not all.

The learning algorithm of a network is closely related to the structure of the network. Since the CSRN is a recursive network, the Kalman filter works well for its training algorithm since the Kalman filter is a recursive filter that estimates that state in a system based off previous measurements [11]. The Kalman computations are briefly summarized as follows. The following variables are used in the Kalman filter calculations [9]:

- w(n) is the state vector of the system, the weights in this case
- C(n) is the measurement matrix, or the Jacobian of current output with respect to weights
- G(n) is the Kalman Gain which determines the correction to the state estimation
- $\Gamma$(n) is a conversion factor that relates the filtered estimated error to the actual error α(n)
- R(n) is the measurement noise
- K(n, n-1) is the error covariance matrix

The equations can be summarized as follows.

$$\Gamma(n) = C(n) K(n, n-1) C^T(n) + R(n) \quad (2.1)$$

$$G(n) = K(n, n-1) C^T(n) \Gamma(n)^{-1} \quad (2.2)$$

$$w(n+1|n) = w(n|n-1) + G(n) \alpha(n) \quad (2.3)$$

$$K(n+1, n) = K(n, n-1) - G(n) C(n) K(n, n-1) \quad (2.4)$$

The basic Kalman equations are used to calculate the updates to the weights in the CSRN. These equations also proved to be a performance bottleneck which is addressed in section 3.3.

2.5 Clustering

Clustering is a form of unsupervised learning that seeks to group similar objects together. By supplying a clustering algorithm with different parameters for several objects, it determines the appropriate group to place each object in. This is done based on two ideas. First, all objects in a group should be as similar as possible. Second, each group should be as different as possible. The way in which the similarity of difference of two objects is determined is by means of a distance measure. The difference in distance measurements will create different clusters. Some of the most common distance measurements include the Euclidean distance, the Hamming distance, and the maximum norm [12]. K-means is one of the most common clustering algorithms and requires one to manually choose the number of clusters to create. The algorithm then randomly generates the chosen number of clusters. Each point is then assigned to the cluster whose centroid is closest. The cluster's centroid is the average (arithmetic mean of each variable) of all the points within the cluster. Next the cluster centroid are recomputed after all of the points have been assigned to a cluster. The points are then reassigned to the new clusters. The calculation of the new cluster centroid and reassignment of points to new clusters is repeated until some convergence criterion is met, usually when the cluster assignments do not change [12].

3. Methods

3.1 Clustering of Mazes

Following the topological clustering approach in [7], a similar clustering was accomplished by dividing the mazes into quadrants. The goal is considered as the reference of origin. It was believed that by using this extra information from the resulting clusters, the network would be better able to correctly solve a given maze. The hypothesis was that one needs to move in the same general direction for cells in each quadrant to reach the goal. However, the goal could be in the corner of the maze which would force the whole maze to group in just one cluster. To eliminate this possibility, the row and column information were also included when clustering. This causes a large quadrant to be split into pieces. One piece of a quadrant might be the half closest to the goal while the second piece might be the half farthest away from the goal. Supervised k-means clustering is used with seven variables in this work. The first two variables are the row and column to help split up large quadrants as previously discussed. The next four variables contain directional information (up, down, left and right). The direction will be assigned a 1 if that direction moves closer to the goal and a -1 if the direction moves further away from the goal. The seventh variable contains information about the type of cell. Path cells are assigned a 0 while obstacles are assigned a 1 and the goal is assigned a -1. Once the clusters are defined, they are sorted based on their centroid's distance to the goal. One variation to basic clustering that is referred to as clustering during epochs. This variation clusters the maze using a slightly different approach. Clustering is performed during the 4$^{th}$ and every 10$^{th}$ epoch instead of beforehand. Because of this, slightly different variables are used for this implementation. It still uses the row and column information but the four directional variables are defined differently. Instead of using a 1 or -1 to define the directions, the value of the four neighbors from the previous epoch are used. Also, the 7$^{th}$ variable is the value of the current cell rather then a number representing the type of cell. The steps for both clustering implementations can be seen below.

1. Determine the 7 variables for each cell in the maze
2. Call k-means function in Matlab using 24 clusters
3. Sort clusters based on their centroid's distance to the goal cell

3.2  Modification of Existing CSRN Code

After looking into how to use the clusters a problem arose. The original CSRN only had two external input nodes, the obstacle and goal as shown in Fig. 5. Therefore, to use the clustered information the obstacle and/or goal would no longer be able to used as one of the external inputs. All attempts at this failed to improve the results. Then it was decided that in order to successfully use any cluster information additional external inputs had to be added to the network. The current CSRN structure in Fig. 5 allows for two external inputs which are used to designate if a particular cell is an obstacle or the goal.

Adding additional external inputs proved to be much easier than expected using the built in ability to add extra recurrent nodes using the publicly available CSRN code [15]. In essence, how each node was identified was simply redefined. Originally the nodes were identified based on the first node. In this scenario, several offset pointers embedded deep within the code would have required changes. However, by redefining the nodes based on the last node in Fig. 5, adding additional external inputs simply required increasing the total number of nodes in the network. Therefore, redefining the nodes based on the last one, were able to add new nodes to the front instead of the back.

3.3 Performance Improvements

*a) Performance Metrics:* The first experiments used the existing form of performance evaluation metric, the "Goodness of Solution" function [3]. The "Goodness of Solution" was created to ascertain what percent of pathway cells point in the correct direction. However, it was discovered that the goodness function did not always indicate a correct calculation. The "Goodness of Solution" function only allows one correct direction. In many cases, there may be two directions that are equally correct. An example of a cell having two correct directions can be seen Fig. 6 by looking at the shaded cell. Therefore, a new performance metric is introduced, called "Correctness of Solution." This function more accurately portrays how well the network has calculated the maze. The correctness of solution considers a cell correct if its neighbor with the lowest value is in the same direction as one of the possible good choices in the target maze. Since the "Correctness of Solution" accounts for multiple good directions while the "Goodness of Solution" only accounts for one good direction, the "Correctness of Solution" will always be greater than or equal to the "Goodness of Solution." Figure 7 simply shows that the "Correctness of Solution" will detect more correct cells than the "Goodness of Solution." The results in Fig. 7 and all following results were obtain using five random 12x12 mazes for training and five different, random 12x12 mazes for testing.

| 4  | 3  | 25 | 5  | 4  |
|----|----|----|----|----|
| 25 | 2  | 1  | 25 | 3  |
| 25 | 1  | 0  | 1  | 2  |
| 5  | 25 | 1  | 2  | 25 |
| 4  | 3  | 2  | 3  | 4  |

*Figure 6: The shaded area is an example of a cell with two correct directions.*

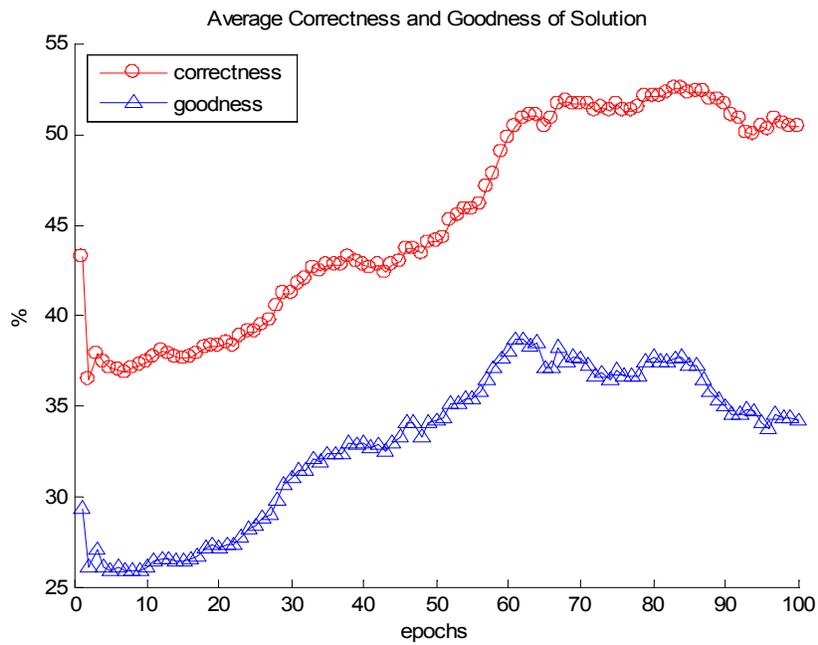

*Figure 7: Plot of goodness and correctness verse epochs.*

*b) Computation Speed:* The CSRN is trained using the extended Kalman filter (EKF) method. This method is used for state estimation in many controls problems. Calculating the Kalman gain, G(n), requires taking the inverse of Γ(n), as given in Eq. (1.2), which is very computationally intensive. This matrix becomes larger as one increases the size of the maze and the number of training samples. Taking the inverse of any matrix is very costly, let alone a large matrix. Review of code optimization techniques in Matlab [13] shows that instead of taking the inverse of Γ(n), one may simply divide the product of the rest of the terms in G(n) by Γ(n). This change is reflected in Eq. (3.1) which is an optimized version of Eq. (2.2).

$$G(n) = K(n, n-1) C^T(n) / \Gamma(n) \quad (3.1)$$

Since division is less computationally intensive, this can make a great improvement when working large mazes and/or many training samples. However, when a few small mazes are used, not much improvement in speed may be expected. Also, it should be noted that multiplying by the inverse and dividing by the matrix do not offer exactly the same result. Due to the rounding of calculations there is a slight difference. For a 12x12 Maze the SSE is less than $10^{-3}$ (1.3231e-004). However, since the matrix is used for state estimation an exact value is not needed.

Another improvement was obtained during the Kalman calculations as well. When updating the Jacobian matrix, C(n), rows were added one at a time. In Matlab, unlike C, it is easy to add a row to a matrix code-wise. However, even though writing code that augments a matrix is easy, the task is still very computationally intense. When a row is added to a matrix, Matlab automatically creates a new matrix of the appropriate size. Then all the data from the old matrix is copied to the new matrix and the old matrix is deleted. This can also lead to memory fragmentation. In the original code, adding one row at a time became a performance bottleneck. To overcome this bottleneck, the code was changed so that several rows were added at once. This provided a noticeable speed improvement as well.

4. Results

4.1 Clustering of Mazes

The first attempts to cluster mazes were an attempt to simulate rooms as in Ref. 7. Figure 8 shows a sample maze used in Ref. 7 that partially inspired this work. This maze consists of rooms in which the majority of the states must travel in the same direction to reach the exit. States are indicated by the arrows.

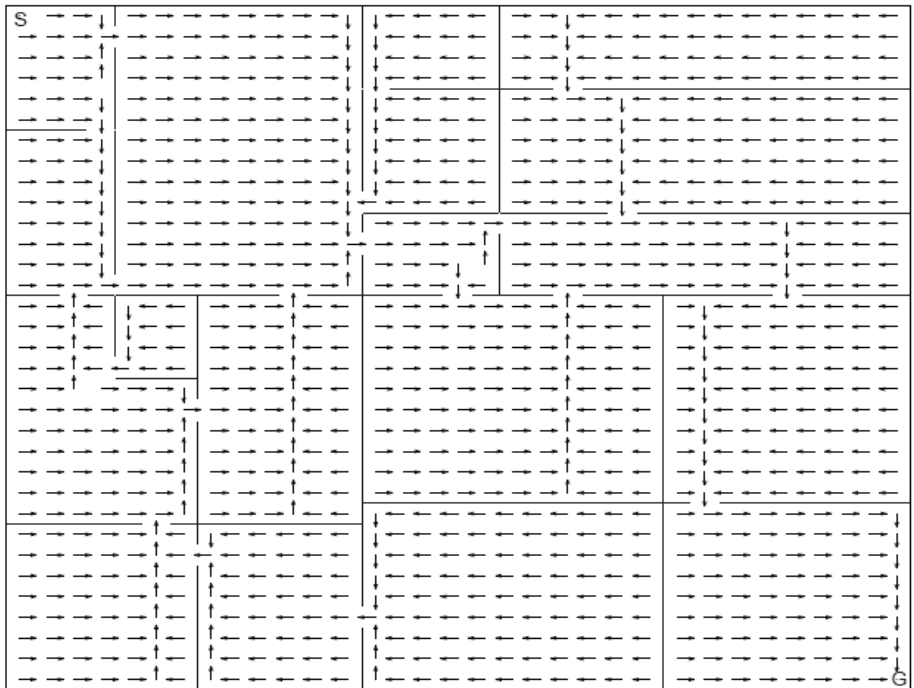

*Figure 8: Influential maze from [7].*

The mazes were successfully clustered in a way that visually appeared similar to the work of [7], as seen in Fig. 10. To understand Fig. 10, note Fig. 9 which shows a graphical representation of the solved maze. The darkest red cell towards the bottom in the maze represents the goal and the random dark blue cells are the obstacles. The colors range from red to yellow to blue in order of increasing distance from the goal. In Fig. 10, one can see that the cells of a particular cluster must generally travel in the same direction to reach the goal. For example, the cells belonging to the cluster in the bottom left corner must all travel up and to the right in order to reach the goal.

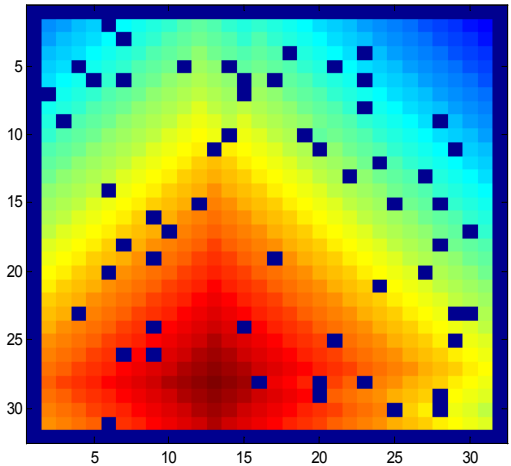

*Figure 9: Graphical Representation of a solved maze.*

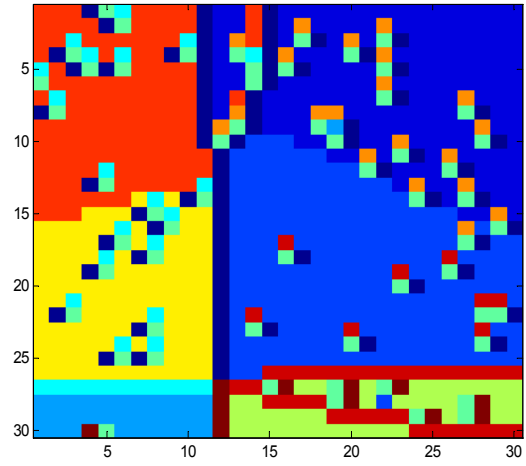

*Figure 10: A clustered version of the maze depicted in Figure 8.*

## 4.2 Modification of Existing CSRN Code

The first attempts to improve the network through clustering used the "Goodness of Solution" measurement to judge the results. This measurement showed little improvement for individual trials using the various methods experimented with. First the cluster number of the cell was added as the third external input to the CSRN. When that failed to show significant improvement for the goodness measurement another method was next tried to cluster the maze based on the values of each cell's neighbors in the middle of training. Training was initiated without any information in the third node but added clustered information during the 4$^{th}$ and every 10$^{th}$ epoch. This clustering during epochs method more closely followed the approach used in [7] and discussed in Section1. Since the clustering during epochs method did not show significant improvement in the goodness for individual trials, next dividing the maze into submazes was tried. A similar technique with this network has been shown to show potential for affine transforms [14]. Since the publicly available CSRN implementation [15] is hard coded to use square matrices, larger mazes were divided the into several smaller square mazes. For example, a 12x12 maze is divided into nine 4x4 submazes as shown in Fig. 11.

| 11 | 10 | 9  | 8 | 7 | 8  | 25 | 10 | 9  | 10 | 11 | 12 |
|----|----|----|---|---|----|----|----|----|----|----|----|
| 10 | 9  | 25 | 7 | 6 | 7  | 8  | 25 | 8  | 25 | 10 | 11 |
| 9  | 8  | 7  | 6 | 5 | 6  | 7  | 6  | 7  | 8  | 9  | 10 |
| 8  | 7  | 6  | 5 | 4 | 5  | 6  | 5  | 6  | 7  | 8  | 9  |
| 7  | 6  | 5  | 4 | 3 | 4  | 25 | 4  | 5  | 25 | 7  | 8  |
| 6  | 5  | 4  | 3 | 2 | 25 | 2  | 3  | 4  | 5  | 6  | 7  |
| 5  | 4  | 3  | 2 | 1 | 0  | 1  | 2  | 3  | 4  | 5  | 25 |
| 6  | 5  | 4  | 3 | 2 | 1  | 2  | 3  | 4  | 25 | 6  | 7  |
| 25 | 6  | 5  | 4 | 3 | 2  | 3  | 4  | 5  | 6  | 7  | 8  |
| 25 | 25 | 6  | 5 | 4 | 3  | 4  | 5  | 25 | 7  | 8  | 9  |
| 9  | 8  | 7  | 6 | 5 | 4  | 5  | 6  | 7  | 8  | 9  | 10 |
| 10 | 9  | 8  | 7 | 6 | 5  | 6  | 7  | 8  | 9  | 10 | 11 |

*Figure 11: 12x12 maze divided into nine 4x4 submazes.*

Once again the results failed to show a great improvement for the individual trials. So, investigation was done to determine what information would be most useful to the network. Since the steps needed to reach the goal is a sort of measurement of the distance to the goal, the Euclidean distance was investigated. The Euclidean distance was added as a third external input to the network and began using the Correctness of Solution metric which produced very good results, as shown in the Fig. 12 using the Correctness of Solution metric. This figure shows that approximately 94% of all the cells in the maze point in the correct direction. This performance is to be expect since the Euclidean distance can be seen as an approximation of the number of steps to the goal. Thus, the network is fed an approximate solution into its third external input. Figure 13 shows the same maze traversing solution results using the Correctness of Solution metric, however, for the original CSRN configuration [15] for comparison.

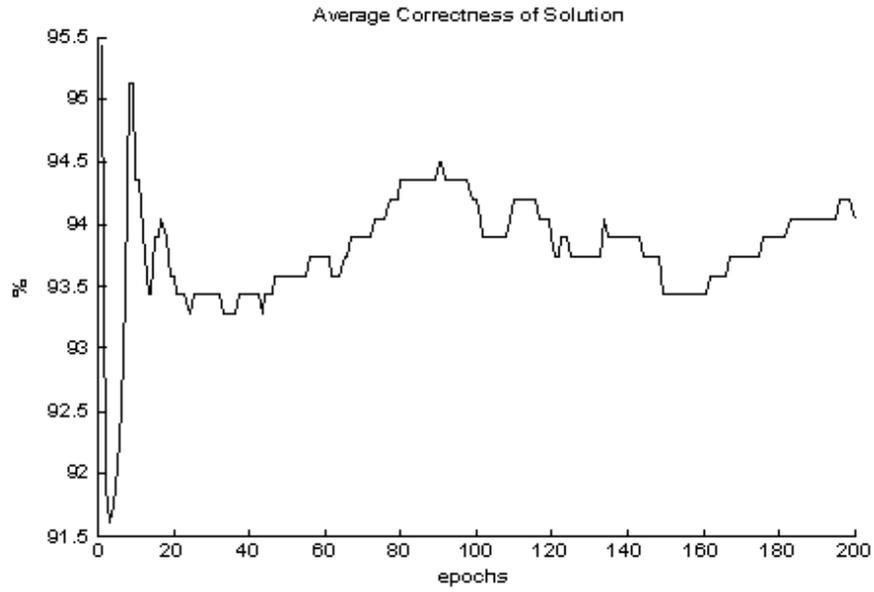

*Figure 12: Results from using the euclidean distance as an external input to the network.*

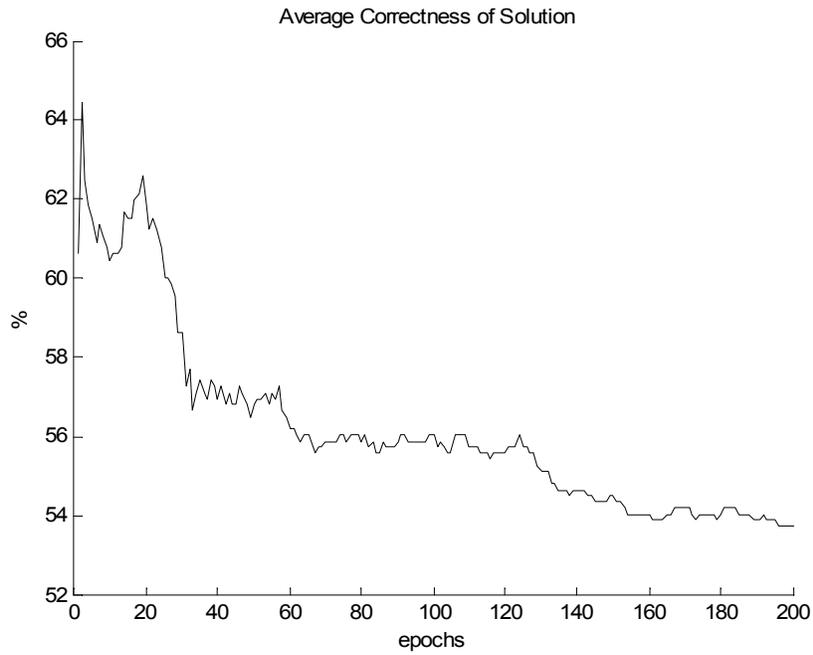

*Figure 13: Results from the original CSRN.*

Despite improved average Correctness of Solution performance from using the Euclidean distance as an additional external input to the CSRN, the initial goal was to cluster the mazes. We then retested the previous methods with the Correctness of Solution metric, discussed in Section 3.3. Interestingly, it was found that when using the Correctness of Solution metric, the original method of using the cluster number as the third external input does improve the performance. The two other methods, clustering during epochs and submazes, showed slightly lower correctness then the original CSRN. It is also worth noting that even though the initial individual trials did not show significant improvements in goodness with any of the methods, the batch results, shown in Table 1, show improved goodness for plain clustering and the submaze methods. This is likely due to variations possible within individual tests. Testing a batch of 100 training and testing sets gives us a much more accurate feel for the performance of each method. Table 1 shows the average results for each method run on a batch of 100 training and testing set. All training and testing sets used 5 random 12x12 mazes for training and testing.

| Method | Correctness | Goodness |
| --- | --- | --- |
| Original CSRN | 67.7% | 42.9% |
| Clustering | 82.9% | 53.2% |
| Clustering during epochs | 64.0% | 42.7% |
| Submazes | 66.2% | 51.5% |
| Euclidean Distance | 97.5% | 66.2% |

*Table 1: Average batch results.*

From Table 1, note that the Euclidean distance method produces the best results with an increase of 29.8% using the Correctness of Solution metric. While using the Euclidean distance as one of the inputs to the CSRN might initially appear cheating, it can be considered a viable method as well. In a real life scenario, one may have access to a GPS system that indicates their current position as well as the position of a goal. From this information the Euclidean distance can be determined but that alone may not indicate the correct path to the goal. Furthermore, the original clustering method using the cluster number as the third external input to the CSRN shows very good performance as well. The clustering during epochs and submaze methods resulted in a slight improvement but not enough to warrant much attention. The improvement in performance using the original clustering method is similar to that using the Euclidean distance. Note the clusters are sorted by their distances of centroids from the goal, they contain information relative to each cell's distance to the goal. Also, note that this clustering approach is different from the earlier hypothesis that clusters moving in the same direction will accelerate the CSRN's maze traversing performance. Therefore, it is observed that as the number of clusters are increased, the performance should also increases. Theoretically, if the number of clusters is increased to the number of cells in the maze then the performance will be the same as the Euclidean distance.

4.3 Performance Improvements

By optimizing the original code for the CSRN, training and testing the network was much quicker. As seen in Fig. 14 and Fig. 15, the first change allowed for an improvement from 257 seconds to just 118 seconds in the Kalman step function. For this run, the time for the Kalman step function was decreased by approximately 54%. This improvement was made in the Kalman step function. The Kalman step function contains the calculations from equations 2.1-2.4. The second improvement is found in the KalmanAddRowToJacobian function. This improvement decreased the time by approximately 70%. Since these two functions took the most time, the overall time for the main function was decreased from 476 to 218 seconds, a decrease of 54% for the entire program. These results were obtained on a laptop with an AMD Athlon 64 X2 (TK-57) processor with 4 GB of RAM.

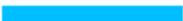

*Figure 14: Run time for Original Code.*

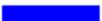

*Figure 15: Run Time for Optimized Code.*

5. Future Works

For future work, a comprehensive analysis of performance will be performed for the number of clusters relative to the size of the maze. Also, using clustering in combination with additional external inputs may improve results further. Detailed analysis of the effects of different parameters may also offer a better understanding of what information will help to improve the network's performance. Furthermore, one may expand this work of clustering to other application domains such as image processing applications.

6. Acknowledgements

The author would like to acknowledge Dr. Khan Iftekharuddin for his guidance and oversight in this work, Dr. David Russomanno, Dr. Robert Kozma, Dr. Salim Bouzerdoum and his research group at the University of Wollongong, and National Science Foundation (NSF) grants MemphiSTEM NSF-DUE 0756738, ECCS-0738519 and ECCS-0715116 for partial support of this work. Also, NSF grant CSEMS NSF-DUE 0422097 and S-STEM NSF-DUE 0630899 provided scholarship support during my undergraduate studies.